\documentstyle[nlprs97,epsbox]{article}
\title{Resolution of Verb Ellipsis in Japanese Sentence\\ using Surface Expressions and Examples}
\author{Masaki Murata, Makoto Nagao \\[0.3cm]
Department of Electronics and Communication,Kyoto University,\\
 Yoshida-honmachi, Sakyo, Kyoto, 606-01, Japan\\
\{mmurata,nagao\}@kuee.kyoto-u.ac.jp\\
tel : 075-753-5962, fax : 075-751-1576}
\date{}

\renewcommand{\topfraction}{1}

\begin{document}
\renewcommand{\topfraction}{1}

\bibliographystyle{fullname}

\maketitle

\begin{abstract}
Verbs are sometimes omitted in Japanese sentences. 
It is necessary 
to recover omitted verbs 
for purposes of language understanding, 
machine translation, and conversational processing. 
This paper describes a practical way to recover omitted verbs
by using {\it surface expressions} and {\it examples}. 
We experimented the resolution of verb ellipses 
by using this information, 
and obtained a recall rate of 73\% and 
a precision rate of 66\% on test sentences. 
\end{abstract}

\section{Introduction}
\label{sec:6c_intro}

Verbs are sometimes omitted in Japanese sentences. 
It is necessary to resolve verb ellipses   
for purposes of language understanding, 
machine translation, and dialogue processing. 
Therefore, we investigated verb ellipsis resolution 
in Japanese sentences. 
In connection with our approach, 
we would like to emphasize the following points: 
\begin{itemize}
\item 
Little work has been done so far 
on resolution of verb ellipsis in Japanese. 

\item 
Although much work on verb ellipsis in English 
has handled 
the reconstruction of the ellipsis structure 
in the case when the omitted verb is given, 
little work has handled 
the estimation of 
what is the omitted verb 
\cite{Dalrymple91} \cite{Kehler93} \cite{Lappin96}. 
On the contrary, 
we handle the estimation of 
what is the omitted verb. 

\item 
In the case of Japanese, 
the omitted verb phrase is sometimes not in the context, 
and the system must construct the omitted verb 
by using knowledge (or common sense). 
We use example-based method to solve this problem. 
\end{itemize}

This paper 
describes a practical method to recover omitted verbs 
by using surface expressions and examples. 
In short, 
(1) when the referent of a verb ellipsis is in the context, 
we use surface expressions (clue words); 
(2) when the referent is not in the context, 
we use examples (linguistic data). 
We define the verb to which a verb ellipsis refers 
as {\it the recovered verb}. 
For example, 
``[KOWASHITA]\footnote{
\label{foot:6c_bracket}
A phrase in brackets ``['',``]'' represents an omitted verb.}
 (broke)'' in the second sentence 
of the following example 
is a verb ellipsis. 
``KOWASHITA (broke)'' in the first sentence 
is a recovered verb. 
\begin{equation}
  \begin{minipage}[h]{10cm}
\small
  \begin{tabular}[t]{lll}
KARE-WA & IRONNA MONO-WO & KOWASHITA.\\
(he)& (several things) & (broke) \\
\multicolumn{3}{l}{
(He broke several things.)}\\
\end{tabular}
  
\vspace{0.3cm}

  \begin{tabular}[t]{lll}
KORE-MO & ARE-MO & [KOWASHITA].\\
(this)  & (that) & (broke) \\
\multicolumn{2}{l}{
([He broke] this and that.)}\\
\end{tabular}
\end{minipage}
\end{equation}

(1) When a recovered verb exists in the context, 
we use surface expressions (clue words). 
This is because 
an elliptical sentence in the case (1) is 
in one of several typical patterns and 
has some clue words. 
For example, 
when the end of an elliptical sentence is the clue word ``MO (also)'', 
the system judges that 
the sentence is a repetition of the previous sentence 
and the recovered verb ellipsis is 
the verb of the previous sentence. 

(2) When a recovered verb is not in the context, 
we use examples. 
The reason is that 
omitted verbs in this case (2) 
are diverse and 
we use examples to construct the omitted verbs. 
The following is an example of 
a recovered verb that does not appear in the context. 
\begin{equation}
  \begin{minipage}[h]{10cm}
    \begin{tabular}[t]{lll}
      SOU & UMAKU IKUTOWA & [OMOENAI] .\\
      (so) & (succeed so well) & (I don't think)\\
\multicolumn{3}{l}{
  ([I don't think] it succeeds so well. )}
    \end{tabular}
  \end{minipage}
\label{eqn:6c_souumaku}
\end{equation}
When we want to resolve the verb ellipsis 
in this sentence ``SOU UMAKU IKUTO WA [OMOENAI]'', 
the system gathers sentences 
containing the expression 
``SOU UMAKU IKUTOWA (it succeeds so well. )'' from corpus 
as shown in Figure~\ref{tab:how_to_use_corpus}, 
and judges that 
the latter part of the highest frequency in the obtained sentence 
(in this case, ``OMOENAI (I don't think)'' etc.) 
is the desired recovered verb. 

\begin{figure}[t]
\small
  \begin{center}
\hspace*{-0.3cm}
  \begin{tabular}[t]{l@{ }l@{ }l}
 & {\bf The matching part} & {\bf The latter part}\\[0.2cm]
KON'NANI & \underline{UMAKU IKUTOWA} & OMOENAI. \\
(like this)  & (it succeeds) & (I don't think)\\
\multicolumn{3}{l}{
(I don't think that it succeeded like this)}\\[0.2cm]
ITUMO & \underline{UMAKU IKUTOWA} & KAGIRANAI.\\
(every time)  & (it succeeds) & (cannot expect to)\\
\multicolumn{3}{l}{
(You cannot expect to succeed every time.)}\\[0.2cm]
KANZENNI & \underline{UMAKU IKUTOWA} & IENAI.\\
(completely)  & (it succeeds) & (it cannot be said)\\
\multicolumn{3}{l}{
(It cannot be said that it succeeds completely)}\\
\end{tabular}
  \end{center}
\caption{Sentences containing ``UMAKU IKUTOWA (it succeeds)'' in a corpus (examples)}
\label{tab:how_to_use_corpus}
\end{figure}

\section{Categories of Verb Ellipsis}

We handle only verb ellipses in the ends of sentences. 


\begin{figure*}[t]
      \begin{center}
      \epsfile{file=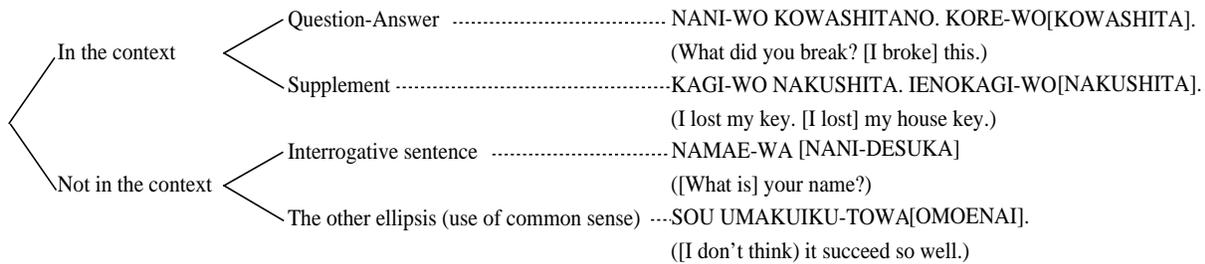,height=3.5cm,width=16cm} 
      \end{center}
    \caption{Categories of verb ellipsis}
    \label{fig:shouryaku_bunrui}
\end{figure*}

We classified verb ellipses 
from the view point of machine processing. 
The classification is shown in Figure \ref{fig:shouryaku_bunrui}. 
First, we classified 
verb ellipses 
by checking 
whether there is a recovered verb in the context or not. 
Next, 
we classified 
verb ellipses by meaning. 
``In the context'' and ``Not in the context'' 
in Figure \ref{fig:shouryaku_bunrui} 
represent where the recovered verb exists, respectively. 
Although 
the above classification is not perfect 
and needs modification, 
we think that it is useful 
to understand the outline of verb ellipses in machine processing. 

The feature and the analysis 
of each category of verb ellipsis are described 
in the following sections. 

\subsection{When a Recovered Verb Ellipsis Appears 
in the Context}

\subsubsection{Question--Answer}

In question--answer sentences  
verbs in answer sentences are often omitted, 
when answer sentences use the same verb 
as question sentences. 
For example, 
the verb of ``KORE WO (this)'' is omitted 
and is ``KOWASHITA (break)'' in the question sentence. 
\begin{equation}
  \begin{minipage}[h]{11.5cm}
  \begin{tabular}[t]{ll}
NANI-WO & KOWASHITANO\\
(what)& (break)  \\
\multicolumn{2}{l}{
(What did you break?)}\\
\end{tabular}
  
\vspace{0.3cm}

  \begin{tabular}[t]{ll}
KORE-WO & [KOWASHITA].\\
(this)  & (break)\\
\multicolumn{2}{l}{
([I broke] this.)}\\
\end{tabular}
\end{minipage}
\end{equation}

The system judges whether 
the sentences are question--answer sentences or not 
by using surface expressions such as ``NANI (what)'', 
and, if so, 
it judges that the recovered verb 
is the verb of the question sentence. 

\subsubsection{Supplement}

In sentences which play a supplementary role to 
the previous sentence, 
verbs are sometimes omitted. 
For example, 
the second sentence 
is supplementary, explaining 
that ``the key I lost'' is ``house key''. 
\begin{equation}
  \begin{minipage}[h]{11.5cm}
  \begin{tabular}[t]{ll}
KAGI-WO & NAKUSHITA.\\
(key)  & (lost)\\
\multicolumn{2}{l}{
(I lost my key.)}
\end{tabular}

\vspace{0.3cm}

  \begin{tabular}[t]{lll}
IE-NO & KAGI-WO & [NAKUSHITA.]\\
(house) & (key)  & (lost)\\
\multicolumn{2}{l}{
([I lost] my house key. )}
\end{tabular}
\end{minipage}
\end{equation}

To solve this, we present the following method 
using word meanings. 
When the word at the end of the elliptical sentence 
is semantically similar to the word 
of the same case element in the previous sentence, 
they correspond, 
and 
the omitted verb is judged to be 
the verb of the word of the same case element in the previous sentence. 
In this case, 
since ``KAGI (key)'' and ``IE-NO KAGI (house key)'' are semantically similar 
in the sense that they are both keys, 
the system judges they correspond,
and the verb of ``IE-NO KAGI-WO (house key)'' is ``NAKUSHITA (lost)''. 

In addition to this method, 
we use methods using surface expressions. 
For example, 
when a sentence has clue words such as the particle ``MO'' 
(which indicates repetition), 
the sentence is judged to be the supplement of the previous sentence. 

There are many cases
when an elliptical sentence is the supplement of the previous sentence. 
In this work, 
if there is no clue, 
the system judges that 
an elliptical sentence is the supplement of the previous sentence. 

\subsection{When a Recovered Verb does not Appear 
in the Context}

\subsubsection{Interrogative Sentence}

Sometimes, in interrogative sentences, 
the particle ``WA'' is at the end of the sentence
and the verb is omitted. 
For example, 
the following sentence is an interrogative sentence 
and the verb is omitted. 
\begin{equation}
  \begin{minipage}[h]{11.5cm}
  \begin{tabular}[t]{ll}
NAMAE-WA & [NANI-DESUKA.]\\
(name)  & (what?)\\
\multicolumn{2}{l}{
([What is] your name?)}
\end{tabular}
\end{minipage}
\end{equation}

If the end is of the form of ``Noun $+$ WA'', 
the sentence is probably an interrogative sentence, 
and thus 
the system judges it to be an interrogative sentence
\footnote{
Since this work is verb ellipsis resolution, 
the system must recover a verb 
such as ``NANI-DESUKA (what is ?)''. 
But the expression of the verb 
changes according to the content of the interrogative sentence 
and we only deal with judging whether the sentence is 
an interrogative sentence or not. }.

\subsubsection{Other Ellipses (Using Common Sense)}

In the case of ``Not in the context'' 
the following example exists 
besides ``Interrogative sentence''. 
\begin{equation}
  \begin{minipage}[h]{11.5cm}
\small
  \begin{tabular}[t]{l@{ }l@{ }l@{ }l}
\footnotesize JITSU-WA & \footnotesize CHOTTO & \footnotesize ONEGAIGA & \footnotesize [ARIMASU].\\
\small (the truth)  & (a little) & (request) & (I have)\\
\multicolumn{4}{l}{
(To tell you the truth, [I have] a request.)}
\end{tabular}
\end{minipage}
\end{equation}
This kind of ellipsis does not have 
the recovered expression in sentences. 
The form of the recovered expression has various types. 
This problem is difficult to analyze. 

To solve this problem, 
we estimate a recovered content 
by using a large amount of linguistic data. 

When Japanese people read the above sentence, 
they naturally recognize the omitted verb is ``ARIMASU (I have)''. 
This is because 
they often use the sentence 
``JITSU-WA CHOTTO ONEGAIGA ARIMASU.
(To tell the truth, I have a request.)'' 
in daily life. 

When we perform the same interpretation using 
a large amount of linguistic data, 
we detect the sentence 
containing an expression 
which is semantically similar to 
``JITSU-WA CHOTTO ONEGAIGA.
(To tell you the truth, (I have) a request.)'', 
and 
the latter part of ``JITSU-WA CHOTTO ONEGAIGA'' 
is judged to be 
the content of the ellipsis. 
To put it concretely, the system detects sentences 
containing the longest characters at the end of the input sentence 
from corpus 
and judges that the verb of the highest frequency 
in the latter part of the detected sentences is 
a recovered verb.

\begin{table*}[t]
  \footnotesize
  \caption{Rule for verb ellipsis resolution}
    \label{tab:doushi_shouryaku_bunrui}
  \begin{center}
\begin{tabular}[c]{|r|p{5cm}|p{3cm}|p{0.5cm}|p{4.8cm}|}\hline
  & Condition & Candidate & \multicolumn{1}{|@{}c@{}|}{Point} & Example sentence\\\hline
\multicolumn{5}{|c|}{Rule in the case that 
a verb ellipsis does not exist}\\\hline
  1 &
  When the end of the sentence is 
  a formal form of a verb 
  or terminal postpositional particles such as 
  ``YO'' and ``NE'', &
  the system judges that a verb ellipsis does not exist. &
  30 & 
  SONO MIZUUMI WA, KITANO KUNINI ATTA.
  (The lake was in a northern country.)
  \\\hline
%
%
\multicolumn{5}{|c|}{Rule in the case of ``Question--Answer''}\\\hline
2&
  When the previous sentence 
  has an interrogative pronoun such as 
  ``DARE (who)'' and ``NANI (what)'', &
  the verb modified by the interrogative pronoun &
  $5$&
  ``DARE-WO KOROSHITANDA'' 
  ``WATASHI-GA KATTE-ITA SARU-WO [KOROSHITA]'' 
  (``Who did you kill?''
  ``[I killed] my monkey'')\\\hline
\multicolumn{5}{|c|}{Rule in the case of ``Supplement''}\\\hline
  3&
  When the end is Noun X followed by a case postpositional particle, 
  there is a Noun Y followed by the same case postpositional particle 
  in the previous sentence, 
  and the semantic similarity between Noun X and Noun Y is 
  a value $s$, &
  the verb modified by Noun Y &
  $s*20$ $-2$&
  SUBETENO AKU-GA NAKUNATTEIRU. 
  GOUTOU-DA-TOKA SAGI-DA-TOKA, ARAYURU HANZAI-GA [NAKUNATTEIRU]. 
  (All the evils have disappeared. 
  All the crimes such as robbery and fraud [have disappeared]. )
  \\
  4&
  When the end is the postpositional particle ``MO'' 
  or 
  there is an expression 
  which indicates repetition such as ``MOTTOMO'', 
  the repetition of the same speaker's previous sentence is interpreted, &
  the verb at the end 
  of the same speaker's previous sentence is judged to be 
  a recovered verb & 
  $5$&
  ``OTONATTE WARUI KOTO BAKARI SHITEIRUNDAYO. 
  YOKU WAKARANAIKEREDO, WAIRO NANTE KOTO-MO [SHITEIRUNDAYO].''
  (``Adults do only bad things. 
  I don't know, but [they do] bribe.'')\\
  5&
  In all cases,&
  the previous sentence&
  $0$&\\\hline
\multicolumn{5}{|c|}{Rule in the case of ``Interrogative sentence''}\\\hline
  6&
  When the end is a noun followed by postpositional particle ``WA'', &
    the sentence is interpreted to be an interrogative sentence.&
  $3$&
  ``NAMAE-WA [NANI-DESUKA]''
  (``[What is] your name?'')
  \\\hline
\multicolumn{5}{|c|}{Rule in the case of use of common sense}\\\hline
  7&
  When the system detects 
  a sentence 
  containing the longest expression at the end of the sentence 
  from corpus, 
  (If the highest frequency is much 
  higher than the second highest frequency, 
  the expression is given 9 points, 
  otherwise it is given 1 point. )
  &
  the expression of the highest frequency 
  in the latter part of the detected sentences 
  &
  1 or 9&
  SOU UMAKU IKUTOWA [OMOENAI]. 
  ([I don't think] it will succeed.)
  \\\hline
\end{tabular}
\end{center}
\end{table*}

\section{Verb Ellipsis Resolution System}

\subsection{Procedure}
\label{6c_wakugumi}

Before the verb ellipsis resolution process, 
sentences are transformed into a case structure 
by the case structure analyzer\cite{csan2_ieice}. 
Verb ellipses are resolved 
by heuristic rules for each sentence from left to right. 
Using these rules, 
our system gives possible recovered verbs some points, 
and it judges that the possible recovered verb 
having the maximum point total is the desired recovered verb. 
This is because a number of types of information is combined 
in ellipsis resolution. 
An increase of the points of a possible recovered verb corresponds to 
an increase of the plausibility of the recovered verb.

The heuristic rules are given in the following form. 
\begin{center}
    \begin{minipage}[c]{12cm}
\small
      \hspace*{0.3cm}{\sl Condition} $\Rightarrow$ \{ {\sl Proposal, Proposal,} .. \}\\
      \hspace*{0.3cm}{\sl Proposal} := ( {\sl Possible recovered verb,}  {\sl Point} )
    \end{minipage}
\end{center}

\noindent
Surface expressions, semantic constraints, 
referential properties, etc., are written as conditions in the {\sl Condition} section. 
A possible recovered verb 
is written in the {\sl Possible recovered verb} section. 
{\sl Point} means the plausibility of the possible recovered verb. 

\subsection{Heuristic Rule}
\label{rule}
\label{sec:6c_ref_pro}

We made 22 heuristic rules for verb ellipsis resolution. 
These rules are made by examining 
training sentences in Section \ref{sec:6c_jikken} by hand. 
We show some of the rules in Table~\ref{tab:doushi_shouryaku_bunrui}. 

The value $s$ in Rule 3 is 
given from the semantic similarity between 
Noun $X$ and Noun $Y$ in EDR concept dictionary \cite{edr_gainen_2.1}. 

The corpus (linguistic data) used in Rule 7 is 
a set of newspapers (one year, about 70,000,000 characters). 
The method detecting similar sentences by character matching 
is performed by sorting the corpus in advance 
and using a binary search.

\begin{figure}[t]
\small

\hspace*{-0.2cm}
\fbox{

\begin{minipage}[h]{7.5cm}

\hspace*{-0.5cm}
  \begin{tabular}[t]{l}
MURI-MO-ARIMASENWA. \\
\multicolumn{1}{l}{
(You may well do so. )}
\end{tabular}

\hspace*{-0.5cm}
  \begin{tabular}[t]{ll}
HAJIMETE & OAISURU-NO-DESUKARA. \\
(for the first time)  & (I meet you)\\
\multicolumn{2}{l}{
(I meet you for the first time)}
\end{tabular}

\hspace*{-0.5cm}
  \begin{tabular}[t]{l@{}l@{}l@{}l}
\footnotesize JITSU-WA & \footnotesize CHOTTO & \footnotesize ONEGAIGA & \footnotesize [ARIMASU].\\
\small (the truth)  & (a little) & (request) & (I have)\\
\multicolumn{4}{l}{
(To tell you the truth, [I have] a request.)}
\end{tabular}


\hspace*{-0.2cm}
\begin{tabular}[h]{|l|r|r|r|}\hline
Candidate  &  the end of  &  ``ARIMASU'' \\
            &  the previous sentence & (I have)\\\hline
Rule 5     &   0 point        &           \\\hline
Rule 7    &              &   1 point     \\\hline
Total score    &   0 point        &   1 point     \\\hline
\end{tabular}


\begin{tabular}[h]{|l|l|}\hline
the latter part of the sentence & Frequency\\
containing ``ONEGAI GA'' & \\\hline
ARIMASU (I have)  & 5\\
ARU     (I have)  & 3\\\hline
\end{tabular}

\caption{Example of verb ellipsis resolution}
\label{tab:6c_dousarei}
\end{minipage}
}\end{figure}

\subsection{Example of Verb Ellipsis Resolution}

We show an example of a verb ellipsis resolution 
in Figure \ref{tab:6c_dousarei}. 
Figure \ref{tab:6c_dousarei} shows that 
the verb ellipsis 
in ``ONEGAI (request)'' 
was analyzed well. 

Since the end of the sentence is not 
an expression which can normally be at the end of a sentence, 
Rule 1 was not satisfied and 
the system judged that 
a verb ellipsis exists. 
By Rule 5 the system took 
the candidate ``the end of the previous sentence''. 
Next, by Rule 7 using corpus, 
the system took the candidate ``ARIMASU (I have)''. 
Although 
there are ``ARU (I have)'' and ``ARIMASU (I have)'', 
the frequency of ``ARIMASU (I have)'' is more than the others 
and it was selected as a candidate. 
The candidate ``ARIMASU (I have)'' 
having the best score was properly 
judged to be the desired recovered verb. 

\section{Experiment and Discussion}

\subsection{Experiment}
\label{sec:6c_jikken}

We ran the experiment on the novel 
``BOKKO- CHAN'' \cite{bokko}. 
This is because 
novels contain various verb ellipses. 
In the experiment, we divided 
the text into 
training sentences and 
test sentences. 
We made heuristic rules by examining training sentences. 
We tested our rules by using test sentences. 
We show the results of verb ellipsis resolution 
in Table \ref{tab:0verb_result}.

\begin{table*}[t]
\fbox{
\begin{minipage}[h]{15.4cm}
    \caption{Result of resolution of verb ellipsis}
    \label{tab:0verb_result}
    \label{tab:6c_sougoukekka}
\vspace*{0.5cm}
  \begin{center}
\begin{tabular}[c]{|lll|rc|rc|rc|rc|}\hline
&&      &\multicolumn{4}{c|}{Training sentences}
        &\multicolumn{4}{c|}{Test sentences}\\\cline{4-11}
&&      &\multicolumn{2}{c|}{Recall}
        &\multicolumn{2}{c|}{Precision}
        &\multicolumn{2}{c|}{Recall}
        &\multicolumn{2}{c|}{Precision}\\\hline
\multicolumn{3}{|l|}{Total}             &  92\% &( 36/41) &  77\% &( 36/47) &  73\% & (33/45) &  66\% & (33/50)\\\hline
 &\multicolumn{2}{l|}{In the context}       & 100\% & (20/20) &  77\% & (20/26) &  85\% & (23/27) &  77\% & (23/30) \\\hline
& &Question--Answer                           & 100\% & ( 3/ 3) & 100\% & ( 3/ 3) & ---\% & ( 0/ 0) & ---\% & ( 0/ 0) \\
& &Supplement                                 & 100\% & (17/17) &  74\% & (17/23) &  85\% & (23/27) &  77\% & (23/30) \\\hline
 &\multicolumn{2}{l|}{Not in the context}  &  76\% & (16/21) &  76\% & (16/21) &  56\% & (10/18) &  50\% & (10/20) \\\hline
& &Interrogative sentence                     & 100\% & ( 3/ 3) &  75\% & ( 3/ 4) & ---\% & ( 0/ 0) &   0\% & ( 0/ 3) \\
& &Other ellipses                             &  72\% & (13/18) &  76\% & (13/17) &  56\% & (10/18) &  59\% & (10/17) \\\hline
\end{tabular}
\end{center}


The training sentences are used to 
make the set of rules in Section~\ref{sec:6c_ref_pro}. \\
{
Training sentences \{the first half of a collection of short stories ``BOKKO CHAN'' \cite{bokko} (2614 sentences, 23 stories)\}\\
Test sentences \{the latter half of novels ``BOKKO CHAN'' \cite{bokko} (2757 sentences, 25 stories)\}

{\it Precision}\, is 
the fraction  of the ends of the sentences 
which were judged to have verb ellipses. 
{\it Recall}\, is the fraction of the ends of the sentences 
which have the verb ellipses. 
The reason why we use precision and recall to evaluate is 
that the system judges that 
the ends of the sentences which do not have 
the verb ellipses 
have the verb ellipses 
and we check these errors properly.
}
\end{minipage}
}
\end{table*}

To judge whether 
the result is correct or not, 
we used the following evaluation criteria. 
When the recovered verb is correct, 
even if the tense, aspect, etc. are incorrect, 
we regard it as correct. 
For ellipses in interrogative sentences, 
if the system estimates that 
the sentence is an interrogative sentence, 
we judge it to be correct. 
When the desired recovered verb appears in the context 
and the recovered verb 
chosen by the rule using corpus 
is nearly equal to the correct verb, 
we judge that it is correct.

\subsection{Discussion}

As in Table \ref{tab:6c_sougoukekka} 
we obtained a recall rate of 73\% and 
a precision rate of 66\% in the estimation of indirect anaphora 
on test sentences. 

The recall rate of ``In the context'' is 
higher than that of ``Not in the context''. 
For ``In the context'' 
the system only specifies the location of 
the recovered verb. 
But 
in the case of ``Not in the context'' 
the system judges that 
the recovered verb does not exist in the context 
and gathers the recovered verb from other information. 
Therefore ``Not in the context'' is very difficult to analyze. 

The accuracy rate of ``Other ellipses (use of common sense)'' 
was not so high. 
But, 
since the analysis of the case of ``Other ellipses (use of common sense)'' 
is very difficult, 
we think that it is valuable 
to obtain a recall rate of 56\% 
and a precision rate 59\%. 
We think that 
when the size of corpus becomes larger, 
this method becomes very important. 
Although we calculate the similarity between 
the input sentence and the example sentence in the corpus 
only by using simple character matching, 
we think that 
we must use the information of semantics and the parts of speech 
when calculating the similarity. 
Moreover 
we must detect the desired sentence 
by using only examples of the type 
(whether it is an interrogative sentence or not) whose previous sentence 
is the same as the previous sentence of the input sentence. 

Although the accuracy rate of the category using surface expressions 
is already high, 
there are some incorrect cases 
which can be corrected 
by refining the use of surface expressions in each rule. 
There is also a case 
which requires a new kind of rule 
in the experiment on test sentences. 
\begin{equation}
  \begin{minipage}[h]{10cm}
\small
  \begin{tabular}[t]{l@{ }l@{ }l@{ }l}
SONOTOTAN & WATASHI-WA & HIMEI-WO & KIITA.\\
(at the moment)& (I)   & (a scream) & (hear)\\
\multicolumn{4}{l}{
(At the moment, I heard a scream?)}\\
\end{tabular}

\vspace{0.2cm}

  \begin{tabular}[t]{l@{ }l@{ }l@{ }l}
{\footnotesize NANIKA-NI} & {\footnotesize TUBUSARERUYOUNA} & {\footnotesize KOE-NO}.\\
(something)& (be crushed)   & (voice)\\
\multicolumn{4}{l}{
\hspace*{-0.2cm}
{\footnotesize (of a fearful voice such that 
he was crushed by something)}}\\
\end{tabular}
\end{minipage}
\end{equation}
In these sentences, 
``OSOROSHII KOE-NO (of a fearful voice)'' 
is the supplement of ``OOKINA HIMEI (a scream)'' in the previous sentence. 
To solve this ellipsis, 
we need the following rule. 
\begin{equation}
  \begin{minipage}[h]{6.5cm} 
    When the end is the form of ``noun X $+$ NO(of)'' and 
    there is a noun Z which is semantically similar to 
    noun Y in the examples of ``noun X $+$ NO(of) $+$ noun Y'', 
    the system judges that 
    the sentence is the supplement of noun Z. 
\end{minipage}
\end{equation}

\section{Conclusion}

In this paper, 
we described a practical way to resolve omitted verbs 
by using surface expressions and examples. 
We obtained a recall rate of 73\% and 
a precision rate of 66\% in the resolution of 
verb ellipsis 
on test sentences. 
The accuracy rate of 
the case of 
recovered verb appearing in the context 
was high. 
The accuracy rate of the case of using corpus (examples) was not so high. 
Since the analysis of this phenomena is very difficult, 
we think that it is valuable 
to have proposed 
a way of solving the problem to a certain extent. 
We think that 
when the size of corpus becomes larger and 
the machine performance becomes greater, 
the method of using corpus will become effective. 

\section*{Acknowledgments}

Research on this work was partially supported by 
JSPS-RFTF96P00502 (The Japan Society for the
Promotion of Science, Research for the Future Program)

{

}
\end{document}